\pdfoutput=1

\documentclass[11pt]{article}
\usepackage{graphicx}
\usepackage{subcaption}
\usepackage{float}
\usepackage[preprint]{acl}
\usepackage{makecell}
\usepackage{multirow}
\usepackage{times}
\usepackage{latexsym}
\usepackage{enumitem}
\usepackage{graphicx}
\usepackage{booktabs}
\usepackage[T1]{fontenc}

\usepackage[utf8]{inputenc}

\usepackage{microtype}

\usepackage{inconsolata}

\usepackage{graphicx}
\usepackage{amsmath}
\usepackage{amssymb}
\usepackage{booktabs}
\usepackage{longtable}
\usepackage{enumitem}
\usepackage{tabularx}
\usepackage{graphicx}
\usepackage{array} 
\usepackage{fancyvrb}
\newcolumntype{l}[1]{>{\arraybackslash}m{#1}}
\title{Enhancing LLM Instruction Following: An Evaluation-Driven Multi-Agentic Workflow for Prompt Instructions Optimization}

\author{Alberto Purpura, Li Wang, Sahil Badyal, Eugenio Beaufrand, Adam Faulkner \\
  Card Intelligence, Capital One \\
  \texttt{\{alberto.purpura, li.wang, sahil.badyal,}\\
	\texttt{eugenio.beaufrand, adam.faulkner\}@capitalone.com}  \\}

\begin{document}
\maketitle
\begin{abstract}
Large Language Models (LLMs) often generate substantively relevant content but fail to adhere to formal constraints, leading to outputs that are conceptually correct but procedurally flawed. 
Traditional prompt refinement approaches focus on rephrasing the description of the primary task an LLM has to perform, neglecting the granular constraints that function as acceptance criteria for its response.
We propose a novel multi-agentic workflow that decouples optimization of the primary task description from its constraints, using quantitative scores as feedback to iteratively rewrite and improve them. Our evaluation demonstrates this method produces revised prompts that yield significantly higher compliance scores from models like Llama 3.1 8B and Mixtral-8x 7B.

\end{abstract}

\section{Introduction}

A key challenge in deploying Large Language Models (LLMs) is ensuring they precisely follow instructions. While LLMs excel at general tasks, they often fail to adhere to specific output constraints, such as word limits, formatting rules or other semantic constraints. This unreliability is a significant barrier to their use in automated systems where strict compliance is non-negotiable, creating a need for better methods to enforce instruction following beyond manual prompt engineering.
\begin{figure}[h!]
\centering
\includegraphics[width=0.75\linewidth]{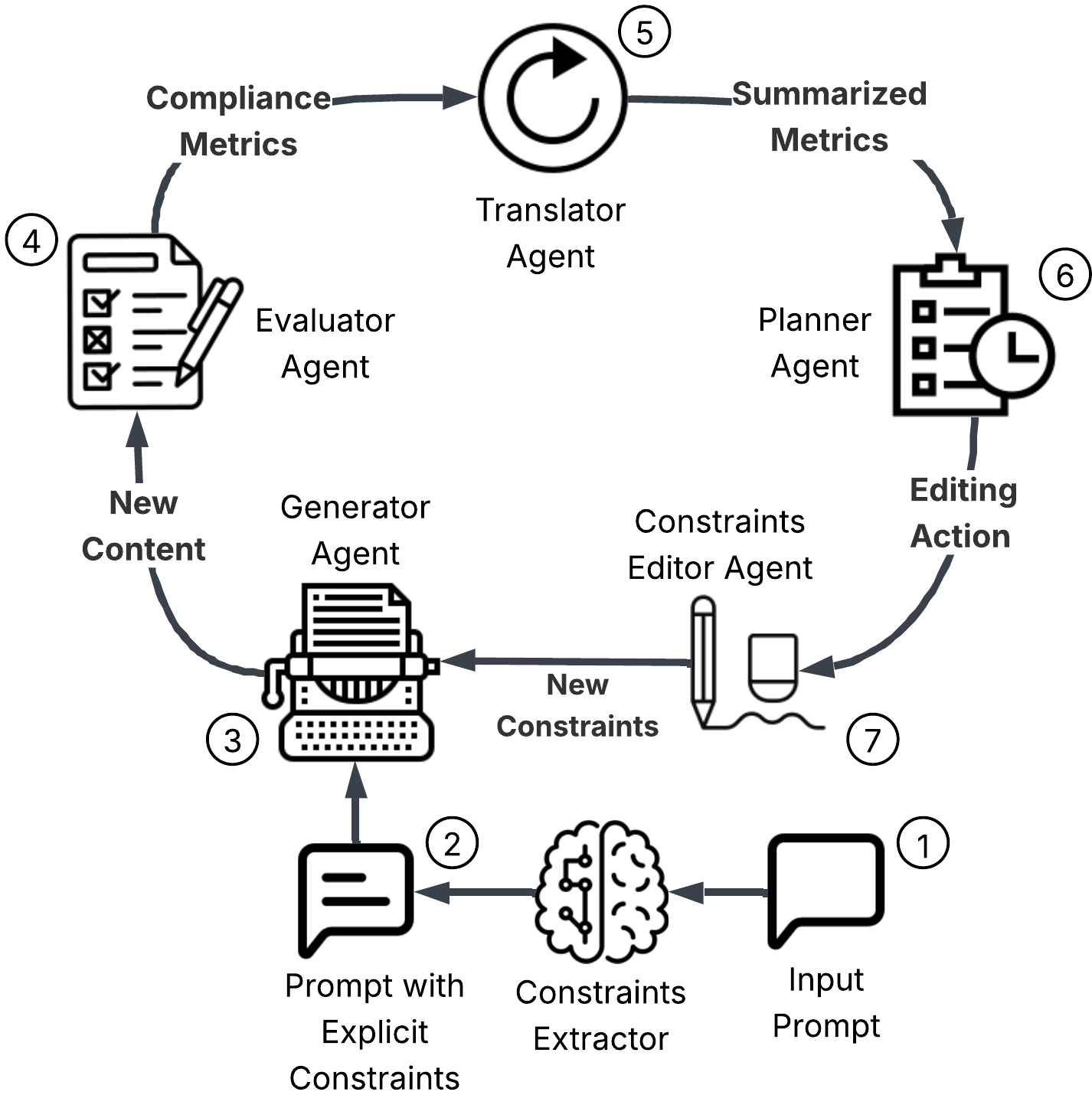}
\caption{Proposed workflow for instruction refinement.}
\label{fig:workflow}
\vspace{-1.5em}
\end{figure}
This paper introduces a novel approach for improving instruction-following within this context by focusing on the explicit decomposition and iterative refinement of response constraints. Our proposed method, depicted in Figure \ref{fig:workflow}, utilizes a dynamic, multi-agentic workflow that evaluates a response's compliance with each specified constraint. Based on this evaluation, the system plans and executes editing actions on the constraints to enhance the quality and compliance of the final output. This approach is distinguished by its modularity and real-world applicability, as it decouples the primary task from the constraints, allowing for targeted refinement aimed at achieving a higher rate of response acceptance based on objective criteria. Finally, note that these acceptance criteria do not only cover form-related constraints but also ones that indicate that the task's goal has been reached.
Our contributions can be summarized as follows:
\begin{itemize}[nosep]
\item We propose and experimentally evaluate the impact of including explicit, modular constraints within a prompt.
\item We introduce a novel, multi-agentic workflow designed to iteratively refine these explicit constraints based on model performance. 
\item We utilize constraint-specific evaluation metrics as direct feedback to guide the refinement process through pre-defined editing actions.
%

\end{itemize}

\section{Related Work}
Work on improving the instruction-following capabilities of LLMs can be classified into three main categories: instruction tuning, representation editing, and prompt engineering.

\noindent\textbf{Instruction tuning} involves fine-tuning LLMs on datasets specifically designed to enhance their ability to follow commands. This is achieved through supervised learning or reinforcement learning using curated or synthesized instruction-response pairs. Representative works like InstructGPT \citep{ouyang2022instructgpt}, Self-Instruct \citep{wang2022selfinstrcut}, COLLIE \citep{yao2023collie}, and WizardLM \citep{xu2024wizardlm} propose novel frameworks for collecting and generating datasets used to improve general instruction-following behavior through fine-tuning.

\noindent\textbf{Representation editing} aims to modify the model’s internal activations during inference to steer text generation toward satisfying given constraints. This is often accomplished by adding a steering vector to the model's hidden states, as explored in works such as \citep{subramani2022steeringvector, zou2023repe, panickssery2023steeringllama2}. Other studies in this area, like \citep{heo2024dollmsknow, stolfo2024activationsteering}, focus on manipulating internal representations to achieve a similar goal. 

\noindent\textbf{Prompt engineering} seeks to improve the quality of the response of an LLM to a prompt without altering model weights or internal states, making it a more accessible and less computationally expensive approach. This category includes methods that enhance an LLM's reasoning and is often related -- although not always equivalent -- to instruction compliance. 
One popular technique is self-refinement on output response, where a model iteratively refines its output response based on feedback from a critic, as seen in  Constitutional AI \citep{bai2022constitutionalai}, Self-Refine \citep{madaan2023selfrefine}, CRITIC \citep{gou2023critic}, and DECRIM \citep{ferraz2024decrim}. However, this approach has shown to potentially underperform prompt rewriting \citep{srivastava2023instances}, an alternative self-refinement approach on input prompt instead, as seen in GRIPS \citep{prasad2022grips}, APE \citep{zhou2022ape}, PromptAgent \citep{wang2023promptagent}, OPRO \citep{yang2023opro} and RaR \citep{deng2023rar}. Nevertheless, these works do not separate constraints from the primary task description in the prompt and rewrite only the constraints. More recently, AIR \citep{liu2025air} presents a method to generate diverse constraints through an iterative process using back-translation and LLM-as-a-Judge. While this capability is notable, for many real-world applications where operational constraints are known beforehand, this approach has little practical value. 
\section{Proposed Approach}
\subsection{Explicit Constraints Extraction} For our evaluation experiments we rely on an extension of the InfoBench dataset \cite{qin2024infobenchevaluatinginstructionfollowing}. We consider each of the original 500 samples in the dataset and report a sample in Table \ref{tab:infobench_sample}. In particular, we consider the columns ``User Input'', ``Instructions'' and ``Decomposed Questions''. The ``User Input'' column contains an (optional) text snippet that, combined with the content of the ``Instructions'' field, constitutes the prompt fed to an LLM. The ``Decomposed Questions'' field is used as part of the evaluation process and contains a series of questions that explain the acceptance criteria of a response to the respective prompt. 
This field was created by the InfoBench authors utilizing an LLM for question generation based on the ``Instructions'' and ``User Input'' fields.
\begin{table*}[h!]
\centering
\small
\resizebox{\linewidth}{!}{%
\begin{tabular}{l{6cm} l{4cm} l{4cm} l{4cm}}
\toprule
\textbf{User Input} & \textbf{Instructions} & \textbf{\makecell{Decomposed \\ Questions}} & \textbf{\makecell{Translated\\Constraints}}\\
\midrule
The typical avocado is over 300 calories from the oil in it. That’s the amount of calories in a large candy bar. If you get enough exercise to eat a large candy bar every day without gaining weight, it wouldn’t be a problem to eat an avocado every day. Other wise you should probably eat them sparingly. & 	
Choose an appealing title for your post. & "Is the generated text a post title?", "Is the generated text appealing as a post tile?", "Is the generated post title suitable for the post in the given input?" & "Ensure the generated text is a post title", "Ensure the generated text is appealing as a post title", "Ensure the generated post title is suitable for the post in the given input"
\\
\bottomrule
\end{tabular}}
\caption{Sample from the dataset we employ in our experiments. The first three fields are sourced from InfoBench \cite{qin2024infobenchevaluatinginstructionfollowing}. The ``Translated Constraints'' column is a new field that we generate, containing the translated version of the ``Decomposed Questions'' so that they may be used as guidelines as part of the prompt fed to an LLM.}
\label{tab:infobench_sample}
\vspace{-1.2em}
\end{table*}
We also add a new column to this dataset, ``Translated Constraints'', that contains a rephrased version of the questions in the ``Decomposed Questions'' field. We use an LLM to rephrase these questions as affirmative statements with the following prompt:
\begin{Verbatim}[fontsize=\small]
You are a question rewriting agent, you will
receive a question from the user and convert it 
to a constraint that instructs to respect the 
given question. For example, if you receive a 
question like 'Is the generated text a post 
title?' you should respond with its 
instructional translation which is 'Ensure 
the generated text is a post title.'
Question: {q}
\end{Verbatim}
Once this process is completed, we create a collection of prompts filling the slots of the following template with the information from the dataset:
\begin{Verbatim}[fontsize=\small]
> System: You are a writing assistant. 
{Instruction}
Ensure your draft complies with all of the 
following requirements: {Translated Constraints}
-- Return only the output required by the task 
and nothing else.
> User: {User Input}
\end{Verbatim}
This process yields a collection of prompts composed of an instruction, an optional user input field and a list of constraints. We also have access to lists of questions corresponding to each of the constraints reported in the prompts. These questions can be used to verify the compliance to each of the constraints using an LLM-as-a-judge approach. While our methodology leverages a pre-existing collection of evaluation questions to generate these constraints, our approach is still generalizable. In fact, the process yielding a collection of questions used to verify each prompt followed by \citet{qin2024infobenchevaluatinginstructionfollowing} can be easily  applied to any existing collection of prompts. In this scenario, an LLM is prompted to decompose the objectives of each prompt into a series of analytical questions. 
\subsection{Prompt Rewriting Workflow}

To enhance instruction-following compliance of a model, we propose a novel, multi-agentic workflow that iteratively refines a modular list of constraints. This system operationalizes three core strategies within an in-context learning framework. 
First, we decouple the primary task indicated in the prompt from the list of constraints the response has to satisfy, allowing the system to focus exclusively on rewriting the instructional components. 
Second, we use constraint-specific evaluation metrics as a direct feedback signal to guide the refinement process. 
Third, we employ a planner-executor architecture to make targeted, intelligent edits to the constraints, to improve the model's compliance.

The workflow, illustrated in Figure~\ref{fig:workflow}, is orchestrated as a stateful graph using LangGraph.
It consists of four primary stages: (i) Content Generation, (ii) Evaluation, (iii) Action Planning, and (iv) Constraint Editing. This cycle repeats until one of the termination conditions is met: a maximum number of iterations, $N_{max}=5$, is reached; the overall response compliance score fails to improve for a predefined number of steps, a threshold we term patience $P_{max}=2$; the response already has a perfect overall compliance score.
We report the prompt templates for each of the agents in Appendix \ref{appendix:workflow_prompts}.

\noindent\textbf{Generator Agent.} Generates a response by following its input prompt. The prompt contains the original Instruction from the dataset, an optional User Input and \textit{current} version of the Translated Constraints as they are edited by the Agents in the workflow. We allow the model to generate three responses to the same prompt and evaluate them in the rest of the workflow.

\noindent\textbf{Evaluator Agent.} Measures a generated responses' fidelity to the prompt's constraints using an LLM-as-a-judge approach. 
This approach evaluates each constraint individually. For a given constraint, its corresponding evaluation question from the InfoBench dataset is presented to a judge model, which assigns a compliance score on a 0-10 scale. We then normalize these scores into the $[0,1]$ interval. The compliance score for a single response is calculated as the mean of its individual constraint scores, and the final metric for a prompt is the average compliance across all generated responses.

To validate this automated evaluation framework, we conducted an inter-annotator agreement study on a subset of 100 prompts. The agreement between two human annotators was \textcolor{black}{96\%}. The agreement between the LLM judge and each human annotator was \textcolor{black}{81\%} and \textcolor{black}{79\%}, respectively. These results demonstrate a relatively strong correlation between human and LLM judgments, thereby supporting the reliability of the approach.


\noindent\textbf{Translator Agent.} Converts the numerical scores from the Evaluator Agent into a textual summary. This summary describes how the constraint compliance score  and the overall response compliance score have changed relative to the previous iteration (e.g., ``increased'', ``decreased'', or ``unchanged''), creating a qualitative feedback report. The summary also contains a history of the responses provided by the Generator Agent and a history of the edits made to the constraints.

\noindent\textbf{Planner Agent.} Serves as the strategic core of the system. Based on the information received from the Translator Agent, the Planner Agent decides which constraint to edit and what action to take, outputting its decision as a detailed instruction. The actions available to this Agent are: (i) ``rephrase'' which changes the wording of one constraint; (ii) ``split'' which divides a long constraint into multiple short ones; (iii) ``merge'' which combines multiple short constraints into a long one (iv) ``reorder'' which changes the order of one or more of the constraints in the prompt. The Agent provides three strategies to edit the prompt and we evaluate them in parallel through the workflow.~\footnote{We include additional experiments on the impact of the chosen number of edit strategies in Appendix \ref{appendix:num_generations}.}

\textbf{Constraint Editing Agent.} Receives the instruction from the Planner Agent and performs the actual modification, rewriting the list of constraints. The resulting new set of constraints is then passed to the Generator Agent in the next iteration.

Upon completion, our workflow returns the version of the constraints associated with the highest overall response compliance score. If the workflow failed to improve the overall compliance score we return the original constraints formulation.
\section{Evaluation Results}
We evaluate our workflow using Llama 3.1 8B and Mixtral-8x 7B as Generator Agents and Llama 3.3 70B for all other Agents. We use greedy decoding for all Agents except for the Planner and Generator Agents where we generate more than one response at each iteration -- in these cases, we set a temperature of 0.9 and rely on nucleus sampling with $top$-$p$ of 0.95 to introduce some variability.

\textbf{Baseline Compliance.} As our first experiment, we compute the compliance of the considered Generator Agent when relying on prompts that do not contain any explicit list of constraints -- i.e., we skip step 2 in Figure \ref{fig:workflow}. In this scenario, we observe an average compliance rate of 81.90\% and 82.59\% for Mixtral-8x 7B and Llama 3.1 8B, respectively. On the other hand, if we include explicit constraints in the prompt -- i.e., if we perform step 2 in Figure \ref{fig:workflow} -- the compliance rates rise to 91.63\% and 91.50\%, respectively. These results confirm the intuition that separating the task description from the response constraints in a prompt increases the compliance rate of its response.

\textbf{Effectiveness of the Multi-Agentic Workflow.} Table \ref{tab:tab_eval} shows the effectiveness of our workflow for improving the response compliance of LLMs.
For prompts that did not already achieve a perfect compliance score, our method successfully increased the score in the majority of cases. In our experiments, we observe an overall increase in response compliance score of 0.0536 and 0.0456 for Mixtral-8x 7B and Llama 3.1 8B, respectively. These scores increase to 0.1306 and 0.1296, respectively if we only consider responses for which we observe an increased compliance. \footnote{We report additional experimental results in Table \ref{tab:compliance_subset} in Appendix \ref{appendix:perf_deltas}, showing the performance increase exact deltas and how they differ when considering prompts from the \textit{Hard} vs \textit{Easy} set from InfoBench.}

\begin{table}[]
\centering
\small
\resizebox{\linewidth}{!}{%
\begin{tabular}{l{1.8cm} l{1.5cm} l{2cm} l{2cm}}
\toprule
\textbf{Generator LLM} & \textbf{Already Compliant (\%)} & \textbf{Unchanged Compliance (\%)} & \textbf{Increased Compliance (\%)} \\
\midrule
Mixtral-8x 7B & 48.49 & 10.46 & 41.05 \\
Llama 3.1 8B & 51.67 & 13.18 & 35.15 \\
\bottomrule
\end{tabular}
}
\caption{Impact of the optimization workflow on prompt compliance, showing the percentage of prompts that were already compliant, unchanged, or improved.}
\label{tab:tab_eval}
\vspace{-1.5em}
\end{table}

\textbf{Validating the Role of Quantitative Feedback.} We also conduct an ablation study to validate the utility of the quantitative compliance scores information for the Planner. When we remove them and force the Planner Agent to rely on just the raw response from the Generator model the rate of positive improvements drops to 38.10\% (-2.95\%) for Mixtral-8x 7B and 34.42\% (-0.72\%) for Llama 3.1 8B. This highlights that direct, quantitative constraint-level feedback is crucial for making effective edits.

\textbf{Actions Distribution Analysis.} We report in Appendix \ref{appendix:planner_actions}, details of the distribution of editing actions selected by the Planner Agent and the number of iterations our workflow performs. The \texttt{rephrase} action is by far the most common strategy, particularly in cases that lead to increased compliance. Successful compliance improvements also tend to require more iterations on average (2.38) compared to cases where compliance remains unchanged (2.00), suggesting that effective refinement is a more persistent process. 
\section{Conclusions}

This paper introduces and experimentally evaluates a novel approach for improving Large Language Model (LLM) instruction-following by incorporating explicit, modular constraints within a prompt. We presented a multi-agentic workflow designed to iteratively refine these constraints to improve the overall compliance of an LLM's response to a prompt. A key differentiator of our approach is its use of constraint-specific evaluation metrics as direct feedback, which guides the refinement process and leads to improved model performance. Our results confirm that this evaluation-driven method of rewriting prompt constraints enhances the compliance of LLM outputs.
We believe this work presents a step toward improving LLM reliability through the granular optimization of instructional constraints, paving the way for their use in automated systems where strict compliance is a hard requirement.
In the future, we aim to extend our evaluation to cover additional evaluation datasets and to experiment with additional models for all the Agents in our workflow.

%

\newpage
\section*{Limitations}
While our proposed multi-agentic workflow demonstrates a promising direction for enhancing LLM compliance, there are some limitations that warrant consideration.

First, our methodology's success is tied to the quality of the initial constraint decomposition. Our experiments leverage the ``Decomposed Questions'' from the InfoBench dataset to generate an initial set of constraints. Although we propose a method for generalizing this approach by using an LLM to generate these questions for any prompt, the effectiveness of the entire optimization cycle is contingent on the quality of this initial decomposition. Flawed or incomplete initial questions could misguide the refinement process, leading to sub-optimal results. In the future, we aim to extend our evaluation to cover additional dataset and strategies for constraints extraction from a prompt.

Second, the evaluation of constraint compliance relies entirely on an LLM-as-a-judge. This approach is subject to the inherent biases and potential inconsistencies of the evaluator model. Furthermore, the iterative and multi-agent architecture, while effective, introduces significant computational overhead and latency. The workflow requires sequential calls to multiple agents, including a large 70B parameter model for planning and evaluation, making it less suitable for real-time or resource-constrained applications. Exploring multiple parallel strategies simultaneously further increases these computational costs.

Finally, the scope of the Planner Agent's actions is confined to a predefined set of operations: rephrasing, splitting, merging, and reordering constraints. This fixed action space may not be sufficient to resolve all types of compliance failures. More complex scenarios might require the ability to introduce new constraints or identify and remove ineffective ones, which is beyond the current capabilities of the system.

\bibliography{custom}

\appendix

\section{Appendix: Additional Experimental Results.}
\label{appendix:perf_deltas}

\begin{table*}[h!]
\centering
\small
\resizebox{\linewidth}{!}{%
\begin{tabular}{
    p{2.2cm} 
    p{1.2cm} 
    >{\centering\arraybackslash}p{1.2cm} 
    >{\centering\arraybackslash}p{1.2cm} 
    >{\centering\arraybackslash}p{2.2cm} 
    >{\centering\arraybackslash}p{2.2cm} 
    >{\centering\arraybackslash}p{2.2cm} 
}
\toprule
\textbf{Generator} & \textbf{Subset} & \textbf{Avg} & \textbf{Stdev} & \textbf{Already} & \textbf{Unchanged} & \textbf{Increased} \\
\textbf{LLM} & & & & \textbf{Compliant (\%)} & \textbf{Compliance (\%)} & \textbf{Compliance (\%)} \\
\midrule
\multirow{2}{*}{Llama 3.1 8B} & Easy & 0.0255 & 0.0886 & 70.92 & 10.76 & 18.33 \\
                          & Hard & 0.0677 & 0.1111 & 30.40 & 15.86 & 53.74 \\
\midrule
\multirow{2}{*}{Mixtral-8x 7B}  & Easy & 0.0257 & 0.0738 & 69.05 & 7.94  & 23.02 \\
                          & Hard & 0.0823 & 0.1358 & 27.35 & 13.06 & 59.59 \\
\bottomrule
\end{tabular}
}
\caption{Breakdown of compliance scores by InfoBench Subset.}
\label{tab:compliance_subset}
\end{table*}
Table \ref{tab:compliance_subset} breaks down the response compliance scores by InfoBench subset difficulty. The results highlight a stark contrast in baseline performance: approximately 70\% of prompts in the \textit{Easy} subset are already compliant, whereas this figure is only about 30\% for the \textit{Hard} subset.

This initial disparity creates a greater opportunity for enhancement among the more challenging prompts. Accordingly, our optimization method led to an increased compliance in over 53\% of cases for the Hard subset, compared to roughly 20\% for the Easy one. This is also reflected in the higher average compliance score increase for the Hard prompts. Interestingly, the percentage of prompts with unchanged compliance remains relatively consistent across both subsets, suggesting that a small portion of prompts are resistant to optimization regardless of difficulty.

\section{Appendix: Validating the Impact of Multiple Editing Strategies.}
\label{appendix:num_generations}
We perform additional experiments investigating whether generating more parallel editing strategies with the Planner Agent improves the response's compliance rate. As shown in Table \ref{tab:eval_different_n_planner}, increasing the number of strategies ($K$) is directly linked to a higher percentage of responses with an increase in the overall compliance metric.
\begin{table}[H]
\centering
\small
\resizebox{\linewidth}{!}{%
\begin{tabular}{l{3cm} l{1cm} l{1cm} l{1cm}}
\toprule
\textbf{Generator LLM} & \textbf{K=1} & \textbf{K=2} & \textbf{K=3}\\
\midrule
Mixtral-8x-7B & 26.05\% & 35.54\% & \textbf{41.05\%}\\
Llama 3.1 8B & 26.16\% & 32.05\% & \textbf{35.15\%}  \\
\bottomrule
\end{tabular}}
\caption{Percentages of responses where we observe an increased compliance score when allowing the Planner Agent to generate K=\{1, 2, 3\} editing strategies.}
\label{tab:eval_different_n_planner}

\end{table}


\section{Appendix: Prompts used by Different Agents in the Proposed Workflow}
\label{appendix:workflow_prompts}

\subsection{Prompt used by the Evaluator Agent.}
\begin{Verbatim}[fontsize=\small]
> System:
You are an expert writing coach acting as a 
fair and strict judge. Your task is to evaluate 
a given passage based on a provided rubric.
> User:
### Description ###
{Task Description}
\% if User Input is not None:
### Input ###
{User Input}
\% endif
### Passage ###
{Response}
### Rubric ###
Evaluate the given passage on the 
following criterion on a scale of 0 to 10:
{Decomposed Question} (0 = completely 
disagree, 2 = somewhat disagree,
5 = neutral, 8 = somewhat agree, 
10 = completely agree).
### Instructions ###
Provide your output only in a JSON format
with the keys "reasoning" and "score".
\end{Verbatim}

\subsection{Prompts used by the Translator Agent.} 
Template for individual constraint scores:
\begin{Verbatim}[fontsize=\small] 
"The compliance score for the constraint \"{}\" 
is {:.2f}, 
{} from the last compliance score {:.2f} by 
{:.2f}."
\end{Verbatim}

\noindent Template for the global average score:
\begin{Verbatim}[fontsize=\small] 
"The average compliance score over all the 
constraints is {:.2f}, {} from the last 
average compliance score {:.2f} by {:.2f}."
\end{Verbatim}

\subsection{Prompt used by the Planner Agent.}
\begin{Verbatim}[fontsize=\small]
> System:
You are an expert prompt reviewer for large 
language models. You have the access to an 
editing history and editing tools. Your task 
is to suggest the best tool to edit the given
 list of constraints such that the average 
 compliance score can be maximized. You should 
understand what has and has not worked 
by reviewing available edits in the 
given history.

> User:
### Editing History ###
{history}

### Editing Tools ###
{
  "rephrase": "change the wording of one 
   constraint to improve its clarity and 
   specificity, without changing the 
   original meaning. This tool is 
   particularly indicated for constraints 
   that have low compliance scores and 
   do not contain examples of how to
   be executed or are not clearly 
   described in an actionable way.", 
  
  "split": "divide a long constraint into 
   multiple short ones to improve its 
   clarity and specificity, without 
   changing the original meaning. 
   This tool is particularly indicated 
   for constraints that are too long and 
   involve multiple actions and have low 
   compliance scores.",
   
  "merge": "combine multiple short 
   constraints into a long one to improve 
   its clarity and specificity, without 
   changing the original meaning.This tool 
   is particularly indicated for constraints 
   that are related or overlapping with each 
   other and both have low compliance 
   scores.",
   
  "reorder": "switch the order of one 
   constraint with the other, without 
   changing the original meaning. This tool 
   is the default whenever the other tools 
   are unlikely to further improve the 
   compliance score of the constraint."
}

### Constraints ###
{constraints}

### Instructions ###
Provide your output only in a JSON format with the 
keys "editing tool" and "how to edit".
\end{Verbatim}


\subsection{Prompt used by the Constraint Editing Agent.}
\begin{Verbatim}[fontsize=\small]
> System:
You are an expert prompt writer for large 
language models. You are given an editing 
suggestion and a list of constraints. 
Your task is to strictly follow the editing 
suggestion and rewrite the list of 
constraints.

> User:
### Editing Suggestion ###
{suggestion}

### Constraints ###
{constraints}

### Instructions ###
Provide your output in the same list format
 of the given constraints without a title.
\end{Verbatim}

\subsection{Examples of Original and Optimized Prompts.}

\textbf{Llama 3.1 8B}.

\begin{itemize}
\item Original prompt 1:
\begin{Verbatim}[fontsize=\small]
> System:
You are a writing assistant. Please 
generate 10 one-sentence hotel reviews 
from ten different customers, ensuring 
that 5 of the reviews are positive and
5 are negative. Begin each review with 
""CUSTOMER"" and the customer's number. 
After completing the 10 reviews, 
provide a two-sentence summarization 
that captures the overall sentiment 
and key points from the reviews.
Ensure your draft complies with all of 
the following requirements: 
-- Ensure the generated text includes hotel 
reviews.
-- Ensure the generated text includes exactly 
10 hotel reviews from 10 different customers.
-- Ensure that each of the generated hotel 
reviews is just one sentence long.
-- Ensure 5 of the reviews in the generated
text are positive and 5 are negative.
-- Ensure that each review in the generated
 text begins with the prefix ""CUSTOMER"" 
 followed by the customer's number.
-- Ensure the generated text includes a 
summarization after completing the 10 reviews.
-- Ensure the summarization in the generated 
text is composed of exactly two sentences.
-- Ensure the generated text captures the 
overall sentiment and key points from the 
reviews in its summarization.
-- Return only the output required by the 
task and nothing else.
> User: 
\end{Verbatim}

\item Optimized prompt 1 (response compliance increase $= 0.1250$): 
\begin{Verbatim}[fontsize=\small]
> System:
You are a writing assistant. Please 
generate 10 one-sentence hotel reviews 
from ten different customers, ensuring 
that 5 of the reviews are positive and
5 are negative. Begin each review with 
""CUSTOMER"" and the customer's number. 
After completing the 10 reviews, 
provide a two-sentence summarization 
that captures the overall sentiment 
and key points from the reviews.
Ensure your draft complies with all of 
the following requirements: 
-- Ensure the generated text includes hotel 
reviews.
-- Ensure the generated text includes exactly
 10 hotel reviews.
-- Ensure the generated text includes reviews
from 10 different customers.
-- Ensure that each of the generated hotel 
reviews is just one sentence long.
-- Ensure 5 of the reviews in the generated 
text are positive and 5 are negative.
-- Ensure that each review in the generated
text begins with the prefix ""CUSTOMER"" 
followed by the customer's number.
-- Ensure the generated text includes a 
summarization.
-- Ensure the summarization is placed after
 completing the 10 reviews.
-- Ensure the summarization in the generated 
text is composed of exactly two sentences.
-- Ensure the summarization captures the 
overall sentiment of the reviews.
-- Ensure the summarization includes the 
key points from the reviews.
-- Return only the output required by the 
task and nothing else.
> User: 
\end{Verbatim}

\item Original prompt 2:

\begin{Verbatim}[fontsize=\small]
> System:
You are a writing assistant. In the 
Board Game Strategy Challenge, 
you are playing a simplified 
version of a strategy board game 
against an opponent. The game 
consists of a 5x5 grid, and you 
have three types of units: Knights 
(K), Archers (A), and Wizards (W). 
Each type of unit has specific 
movement and attack patterns. 
Your objective is to eliminate all 
of your opponent's units. Given 
the initial grid state in the input,
your units are on the bottom row 
(W A K A W), and your opponent's 
units are on the top row 
(A K W A K). Your goal is to design 
specific movement and attack 
patterns for each type of unit,
and then, based on the current 
grid state and rules you designed, 
describe your next move and show 
the grid after your move.
Ensure your draft complies with 
all of the following requirements:
-- Ensure the generated text provides
 a specific design for the movement 
 and attack patterns for each type 
 of unit.
-- Ensure the generated text describes
 the next move.
-- Ensure the next move is based on
the current grid state and the rules 
designed in the generated text.
-- Ensure the generated text 
illustrates an updated grid.
-- Ensure the updated grid is correct
after implementing the described move, 
adhering to the current grid state 
and the rules designed.
-- Return only the output required by 
the task and nothing else.
> User: A K W A K
# # # # #
# # # # #
# # # # #
W A K A W

\end{Verbatim}
\item Optimized prompt 2 (response compliance increase $=0.4800$):

\begin{Verbatim}[fontsize=\small]
> System:
You are a writing assistant. In the 
Board Game Strategy Challenge, 
you are playing a simplified 
version of a strategy board game 
against an opponent. The game 
consists of a 5x5 grid, and you 
have three types of units: Knights 
(K), Archers (A), and Wizards (W). 
Each type of unit has specific 
movement and attack patterns. 
Your objective is to eliminate all 
of your opponent's units. Given 
the initial grid state in the input,
your units are on the bottom row 
(W A K A W), and your opponent's 
units are on the top row 
(A K W A K). Your goal is to design 
specific movement and attack 
patterns for each type of unit,
and then, based on the current 
grid state and rules you designed, 
describe your next move and show 
the grid after your move.
Ensure your draft complies with 
all of the following requirements:
-- Ensure the generated text provides
a specific design for the movement 
and attack patterns for each type of 
unit.
-- Ensure the generated text describes 
the next move.
-- Ensure the next move is based on the 
current grid state.
-- Ensure the next move adheres to the 
rules designed in the generated text.
-- Ensure the generated text includes 
a visual representation of the grid 
after the described move, showing the
 new positions of all units and any 
 changes to the grid state.
-- Ensure the updated grid accurately
 reflects the new positions of all 
 units after the described move, 
 following the movement and attack 
 rules for each unit type and the 
 current state of the grid.
-- Return only the output required 
by the task and nothing else.
> User: A K W A K
# # # # #
# # # # #
# # # # #
W A K A W
"
\end{Verbatim}

\item Original prompt 3:
\begin{Verbatim}[fontsize=\small]
> System:
You are a writing assistant. Design 
a simple training challenge on 
Strava that everyone could attend.
Ensure your draft complies with 
all of the following requirements: 
-- Ensure the generated text is 
a training challenge on Strava.
-- Ensure the training challenge 
is simple enough for everyone to 
attend.
-- Return only the output 
required by the task and nothing 
else.
> User: 
\end{Verbatim}
\item Optimized prompt 3 (response compliance increase $=0.0500$):

\begin{Verbatim}[fontsize=\small]
> System:
You are a writing assistant. Design 
a simple training challenge on 
Strava that everyone could attend.
Ensure your draft complies with 
all of the following requirements: 
-- Ensure the generated text is 
a training challenge on Strava.
-- Ensure the training challenge 
has a manageable duration, 
accessible activity type, and 
minimal requirements, making it 
easy for a wide range of 
participants to join.
-- Return only the output 
required by the task and nothing 
else.
> User: 
\end{Verbatim}
\end{itemize}

\textbf{Mixtral-8x 7B}.
\begin{itemize}
\item Original prompt 1:
\begin{Verbatim}[fontsize=\small]
> System:
You are a writing assistant. 
Choosing a name for your product 
or business YouTube channel is 
an important part of the process.
Based on the description of the
product or business, you should
come up with some interesting 
names. Take some time to 
brainstorm your ideas.
Ensure your draft complies with
all of the following requirements: 
-- Ensure the generated text 
includes some names for a product
or business YouTube channel.
-- Ensure the generated names 
are interesting.
-- Ensure the generated names 
are based on the description of
the product or business in the
given input.
-- Return only the output required 
by the task and nothing else.
> User: Here you will find videos 
and content that will help students
prepare for the application process
to graduate schools as well as how
to apply to graduate schools

\end{Verbatim}

\item Optimized prompt 1 (response compliance increase $=0.0300$):
\begin{Verbatim}[fontsize=\small]
You are a writing assistant. 
Choosing a name for your product 
or business YouTube channel is 
an important part of the process.
Based on the description of the
product or business, you should
come up with some interesting 
names. Take some time to 
brainstorm your ideas.
Ensure your draft complies with
all of the following requirements: 
-- Ensure the generated text 
includes some names for a product
or business YouTube channel.
-- Ensure the generated names
are creative, unique, and memorable.
-- Ensure the generated names are 
based on the description of the 
product or business in the given 
input.
-- Return only the output required
 by the task and nothing else.
> User: Here you will find videos 
and content that will help students
prepare for the application process 
to graduate schools as well as how 
to apply to graduate schools
\end{Verbatim}

\item Original prompt 2:
\begin{Verbatim}[fontsize=\small]
> System:
You are a writing assistant. 
Describe the experience of learning
a new language by drawing parallels
to various stages of a long, 
adventurous journey. Use analogies
to depict at least three challenges
and three milestones one might 
encounter along the way. Your
response should use odd-numbered
sentences and contain the 
longest-length sentence at the 
exact midpoint of the paragraph.
Ensure your draft complies with 
all of the following requirements: 
-- Ensure the generated text is a 
description about the experience of 
learning a new language.
-- Ensure the generated description 
uses analogies to depict the 
experiences of learning a new language.
-- Ensure the generated description 
draws parallels to various stages of 
a long, adventurous journey while 
depicting the challenges and 
milestones one might encounter 
when learning new languages.
-- Ensure the generated description 
includes at least three challenges 
and three milestones one might 
encounter along the way of learning 
a new language.
-- Ensure the generated text uses only 
odd-numbered sentences.
-- Ensure the generated text contains 
the longest-length sentence at the exact 
midpoint of the paragraph.
-- Return only the output required by 
the task and nothing else.
> User: 
\end{Verbatim}
\item Optimized prompt 2 (response compliance increase $=0.3667$):
\begin{Verbatim}[fontsize=\small]
> System:
You are a writing assistant. 
Describe the experience of learning
a new language by drawing parallels
to various stages of a long, 
adventurous journey. Use analogies
to depict at least three challenges
and three milestones one might 
encounter along the way. Your
response should use odd-numbered
sentences and contain the 
longest-length sentence at the 
exact midpoint of the paragraph.
Ensure your draft complies with 
all of the following requirements: 
-- Ensure the generated text is a 
description about the experience of 
learning a new language.
-- Ensure the generated description 
uses analogies to depict the 
experiences of learning a new 
language.
-- Ensure the generated description 
draws parallels to various stages of 
a long, adventurous journey while 
depicting the challenges and 
milestones one might encounter when 
learning new languages.
-- Ensure the generated description 
includes at least three challenges 
and three milestones one might 
encounter along the way of learning 
a new language.
-- Ensure the generated text only 
includes sentences numbered as odd 
integers, in ascending order, 
starting from 1, and does not 
include any even-numbered sentences.
-- Ensure the generated text has its 
longest sentence exactly in the 
middle of the paragraph, with an 
equal number of sentences before and 
after it.
-- Return only the output required by 
the task and nothing else.
> User: 
\end{Verbatim}

\item Original prompt 3:
\begin{Verbatim}[fontsize=\small]
> System:
You are a writing assistant. Solve the 
given Sudoku puzzle.
Ensure your draft complies with all 
of the 
following requirements: 
-- Ensure the generated text is a 
solved Sudoku puzzle with all numbers 
filled in and no blank grids.
-- Ensure the generated text attempts 
to solve the Sudoku puzzle in the given 
input by filling in the missing numbers, 
while keeping the existing filled grids 
in the input unchanged and copying them 
to the generated text without any 
modification.
-- Ensure the generated text is a correct 
solution to the sudoku puzzle in the 
given input.
-- Return only the output required by 
the task and nothing else.
> User: The Sudoku puzzle is:
|1|  |5|  |  |  |  |8|  |
|  |  |  |  |  |3|  |6|4|
|8|3|4|6|7|  |9|  |  |
|2|9|1|  |  |8|3|7|6|
|  |  |6|  |  |9|8|1|  |
|  |  |  |  |  |2|  |  |  |
|  |2|  |  |9|7|6|  |  |
|5|4|  |  |  |  |  |9|8|
|6|1|  |  |  |5|4|3|7|

\end{Verbatim}
\item Optimized prompt 3 (response compliance increase $=0.1333$):

\begin{Verbatim}[fontsize=\small]
> System:
You are a writing assistant. Solve the 
given Sudoku puzzle.
Ensure your draft complies with all 
of the 
following requirements:  
-- Ensure the generated text is a 
complete Sudoku puzzle solution 
where every row, column, and 3x3 
sub-grid contains the numbers 1-9 
without repetition or blank spaces.
-- Ensure the generated text 
attempts to solve the Sudoku 
puzzle in the given input by 
filling in the missing numbers.
-- Ensure the generated text keeps 
the existing filled grids in the 
input unchanged and copies them to 
the generated text without any 
modification.
-- Ensure the generated text is a 
correct solution to the Sudoku puzzle 
in the given input.
-- Return only the output required 
by the task 
and nothing else.
> User: The Sudoku puzzle is:
|1|  |5|  |  |  |  |8|  |
|  |  |  |  |  |3|  |6|4|
|8|3|4|6|7|  |9|  |  |
|2|9|1|  |  |8|3|7|6|
|  |  |6|  |  |9|8|1|  |
|  |  |  |  |  |2|  |  |  |
|  |2|  |  |9|7|6|  |  |
|5|4|  |  |  |  |  |9|8|
|6|1|  |  |  |5|4|3|7|
\end{Verbatim}
\end{itemize}

\section{Appendix: Planner Agent Behavior and Workflow Dynamics}
\label{appendix:planner_actions}
To provide deeper insight into the behavior of our multi-agentic workflow, we analyzed the number of iterations required to achieve compliance improvements and the distribution of editing actions chosen by the Planner Agent. The results, broken down by compliance outcome and the generator model used, are detailed below.

\textbf{Workflow Iteration Analysis.} 
\begin{table}[h!]
\centering
\resizebox{\linewidth}{!}{%
\begin{tabular}{l{6cm}m{4cm}}
\toprule
\textbf{Category} & \textbf{Avg. Number of Cycles} \\ \midrule
Overall & 1.14 \\
Already Compliant &  0.00 \\
Unchanged Compliance  & 2.00 \\
Increased Compliance &  2.38 \\ \midrule
Llama-8b & 1.10 \\
Mixtral-8x7b & 1.18 \\ \midrule
Llama-8b + Already Compliant & 0.00 \\
Llama-8b + Unchanged Compliance & 2.00 \\
Llama-8b + Increased Compliance & 2.39 \\ \midrule
Mixtral-8x7b + Already Compliant & 0.00 \\
Mixtral-8x7b + Unchanged Compliance & 2.00 \\
Mixtral-8x7b + Increased Compliance & 2.37 \\ \bottomrule
\end{tabular}}
\caption{Average number of cycle iterations, categorized by compliance outcome and generator model.}
\label{tab:cycle_iterations}
\end{table}
Table \ref{tab:cycle_iterations} presents the average number of cycle iterations performed by the workflow. As expected, prompts that were already fully compliant required zero iterations. For prompts where compliance was not perfect, the workflow averaged 1.14 iterations overall. Notably, cases resulting in \textit{Increased Compliance} required more iterations on average (2.38) than those with \textit{Unchanged Compliance} (2.00), suggesting that successful optimization often involves a more persistent refinement process. This pattern holds true for both \texttt{Llama-8b} and \texttt{Mixtral-8x7b} models.

\textbf{Planner Action Distribution.} Table \ref{tab:action_stats} details the mean frequency of editing actions selected by the Planner Agent. The \texttt{rephrase} action was overwhelmingly the most common strategy, with an overall mean of 0.88 uses per cycle. This action was particularly prevalent in cases of \textit{Increased Compliance}, where its mean usage rose to 1.81. The \texttt{split} and \texttt{reorder} actions were used more sparingly, while the \texttt{merge} action was never selected, indicating that combining constraints was not identified as a useful optimization strategy in our experiments. We suspect this is due to the fact that in the prompt for the planner agent we suggest the model to use this action when different constraints overlap, a situation that did not occur in the chosen dataset according to the model interpretation of its constraints lists. The data also shows that both \texttt{rephrase} and \texttt{split} actions are more frequently employed when the workflow successfully increases a prompt's compliance score, highlighting their importance in effective constraint refinement.
\begin{table}[h!]
\centering
\resizebox{\linewidth}{!}{%
\begin{tabular}{l{6cm}l{1.5cm}l{1cm}l{1cm}l{1cm}}
\hline
\textbf{Category} & \textbf{Rephrase} & \textbf{Split} & \textbf{Merge} & \textbf{Reorder} \\ \midrule
Overall & 0.88 & 0.14 & 0.00 & 0.12 \\
Already Compliant  & 0.00 & 0.00 & 0.00 & 0.00 \\
Unchanged Compliance & 1.63 & 0.10 & 0.00 & 0.26 \\
Increased Compliance  & 1.81 & 0.35 & 0.00 & 0.22 \\ \midrule
Llama-8b  & 0.80 & 0.19 & 0.00 & 0.12 \\
Mixtral-8x7b  & 0.97 & 0.10 & 0.00 & 0.12 \\ \midrule
Llama-8b + Already Compliant  & 0.00 & 0.00 & 0.00 & 0.00 \\
Llama-8b + Unchanged Compliance  & 1.52 & 0.16 & 0.00 & 0.32 \\
Llama-8b + Increased Compliance & 1.70 & 0.48 & 0.00 & 0.21 \\ \midrule
Mixtral-8x7b + Already Compliant  & 0.00 & 0.00 & 0.00 & 0.00 \\
Mixtral-8x7b + Unchanged Compliance  & 1.77 & 0.04 & 0.00 & 0.19 \\
Mixtral-8x7b + Increased Compliance  & 1.90 & 0.24 & 0.00 & 0.24 \\ \bottomrule
\end{tabular}}
\caption{Mean frequency of planner actions per cycle, categorized by compliance outcome and generator model.}
\label{tab:action_stats}
\end{table}

\end{document}